\documentclass[runningheads]{llncs}

\usepackage{eccv}

\usepackage{eccvabbrv}

\usepackage{graphicx}
\usepackage{booktabs}
\usepackage{multirow}
\usepackage{booktabs}
\usepackage{mathtools}
\usepackage{enumitem}

\usepackage[accsupp]{axessibility}  

\usepackage{hyperref}

\usepackage{orcidlink}

\begin{document}

\title{SWITi: Quantifying and Reducing Tiling Artifacts with Sliding Window Inner Tiling} 

\titlerunning{SWITi}

\author{Federico Carrara\inst{1,2}\orcidlink{0009-0005-9445-8497} \and
Aman Kukde\inst{1}\orcidlink{0009-0008-3955-8382} \and
Melisande Croft\inst{1}\orcidlink{0009-0004-8184-683X} \and
Joran Deschamps\inst{1}\orcidlink{0000-0001-8462-2883} \and
Florian Jug\inst{1}\orcidlink{0000-0002-8499-5812}
}

\authorrunning{F.~Carrara et al.}

\institute{
Fondazione Human Technopole, Milan, Italy \and
Università Campus Bio-Medico, Rome, Italy
}

\maketitle

\begin{abstract}
SWITi is a test-time method for reducing artifacts in tiled predictions, particularly for neural networks that learn posterior distributions from which solutions are sampled at inference time. 
Tiled predictions are unavoidable for large image data, and artifacts arise whenever tiles are smaller than a network's receptive field and when tiles are independent posterior samples. 
SWITi averages overlapping sliding-window predictions, so discrepancies between neighboring samples are spread across shifted tile positions rather than accumulating at fixed seam coordinates. 
For posterior models, SWITi uses no more tile samples than an MMSE estimate requires and therefore incurs no additional forward passes. 
Additionally, we introduce two reference-free metrics, the Fraction of Rejected Tests (FRT) and Artifact Severity (ASV), for detecting and quantifying tiling artifacts from a per-tile permutation test that compares the distribution of pixel gradients across tile seams against the surrounding image content.
On pre-trained and published image splitting models across three fluorescence microscopy datasets in 2D and 3D, we show that SWITi substantially attenuates stitching seams while also improving reconstruction fidelity and resolution. 
Since tiling artifacts in posterior predictions can easily be mistaken for biological structures or for boundaries between biological structures, removing or reducing them using SWITi will improve the downstream processing of large image predictions, which is particularly relevant for biomedical data.
\end{abstract}

\section{Introduction}
\label{sec:intro}
Image data in general, and biomedical images in particular, are routinely larger than what a neural network can process in a single forward pass.
In such cases, we are accustomed to employing tiled training and inference strategies, where the image is partitioned into a regular grid of potentially overlapping tiles, and each tile is then processed independently.
At inference time, the individually predicted tiles are then stitched back into a full-frame output~\cite{ronneberger2015u, isensee2021nnu, stringer2021cellpose, Ashesh2022-ah}.

Consider first the case of deterministic networks with a known receptive field (RF) size.
Wherever a pixel is predicted with its full RF available, its output value is determined by the image content alone.
Two tiles that both cover a pixel with a complete RF thus return the identical prediction, and stitching them is artifact-free by construction.
Hence, adding context to each tile that covers $\frac{1}{2}$ of the RF size on all sides and later discarding that border region when the remaining tiles are placed next to each other avoids tiling artifacts~\cite{Possolo2021-ha, Rumberger2021-pn}.

This strategy, known as \textit{outer tiling}, presumes that the enlarged tile still fits in memory~\cite{Ashesh2022-ah}.
As architectures grew deeper, their RF grew with them, and tiles that still fit into memory became smaller than the RF itself. 
When an output pixel does not have access to the full RF, shifting the input tile location changes the surrounding context, resulting in different predictions for two tiles that cover the same pixel.
A partial remedy is \textit{inner tiling}~\cite{Ashesh2022-ah}, which predicts on tiles of the training size and keeps only a central crop; it attenuates the artifacts but cannot remove them, as the retained inner tile lacks the context of a complete RF~\cite{Ashesh2022-ah, Buglakova2025-no}.

A second independent source of artifacts arises with posterior models, which do not predict a single deterministic output but learn an explicit or implicit posterior $p(y\mid x)$ from which reconstructions are sampled at inference time. 
These models are now widely used for denoising, super-resolution, virtual staining, and image splitting~\cite{Prakash2021-zj, Prakash2021-qx, Salmon2023-vm, Pan2023-tk, Zhang2025-bs, della-Maggiora2023-sq, Ray2025-am, Ray2025-kg, Ashesh2026-da, Carrara2026-fb}.
Here, even a full RF is not enough: independently sampled neighboring tiles need not (and typically will not) agree at their shared boundary.

Neither source can be removed by tiling geometry once the receptive field outgrows the tile size that fits in memory.
Inner tiling leaves a residual RF mismatch, since the retained crop still lacks a complete RF; and a finite sample budget means that averaging independent posterior draws may still leave disagreement across the tile boundary (seam).

These artifacts, however, can be considerably reduced by the method we propose here: \textbf{S}liding \textbf{W}indow \textbf{I}nner \textbf{Ti}ling (SWITi), a tiled-inference strategy that addresses the problems outlined above.
As in inner tiling, each tile contributes only its central region.
Unlike standard minimum mean squared error (MMSE) inner tiling~\cite{Ashesh2022-ah}, which samples the same fixed partition repeatedly and averages the draws in place to approximate the posterior mean, SWITi slides the tile window across the image at a fixed stride and forms the MMSE estimate by averaging the overlapping inner regions.
The stride sets how many inner regions cover each pixel; hence, the number of samples averaged there to compute an approximate MMSE solution.
Inner tiling suppresses the deterministic mismatch as far as the halo allows, and sliding the window keeps the stochastic and receptive field mismatch from accumulating at the fixed tile boundaries.
Note that, at a matched number of samples per pixel, SWITi uses no more forward passes than standard MMSE inner tiling. 
Since it operates entirely at test-time, it applies to any pre-trained model without retraining or architectural change.

Since the removal or reduction of artifacts can only be claimed if they can be detected and quantified, we also introduce a statistical test to detect and measure the severity of tiling artifacts.
No established measure, such as PSNR~\cite{huynhthu2008psnr} or SSIM variations~\cite{wang2004ssim,wang2003msssim,hore2010image,Ashesh2024-ev}, is tailored to this task, and sensitivity is, therefore, as we show, rather weak.
Some dedicated metrics either need the individual tiles before stitching~\cite{Buglakova2025-no} or a ground-truth and non-tiled reference~\cite{wang2003msssim, Ashesh2024-ev, Reina2020-zz}, both of which are typically unavailable when tiled predictions are needed.

We therefore introduce a reference-free, per-tile measure.
For each tile's inner region, we test whether the distribution of pixel gradients across its seams is statistically distinguishable from the distribution in the surrounding image content.
The construction adapts reference-free blockiness metrics from JPEG image-quality assessment~\cite{Wang2003-kk, Perra2005-bh, Liu2009-yz} to inference tiles of arbitrary size, reframed as a non-parametric two-sample test.
Judging each seam against its own local context absorbs content-dependent gradient variation into the control.
Pooling the per-tile rejections yields a detection score, the Fraction of Rejected Tests (FRT), and pooling the test statistics yields a severity score, Artifact Severity (ASV), indicating how pronounced tiling artifacts are in a given result.

While our method is applicable to any network, we evaluate SWITi on image splitting tasks, \ie, the task of recovering individual fluorescent structures from a single superimposed acquisition.
We do this using pre-trained checkpoints of a published and openly available method called Micro$\mathbb{S}$plit~\cite{Ashesh2026-da}, applied in 2D and 3D to three of its most artifact-prone datasets. 
We first validate the test on standard inner tiling predictions, which have so far led to SOTA results, and then show that SWITi substantially attenuates the artifacts we quantify with existing metrics and the metric we introduce here.

\section{Related Work}
\label{sec:rel_work}

\paragraph{\textbf{Tiled inference for dense prediction.}} 
Large biomedical images that do not fit in memory are processed tile-by-tile, with per-tile predictions recombined into a full-frame output. 
The standard recipe, established by Ronneberger et al.~\cite{ronneberger2015u}, is overlap-tile prediction with halo cropping: tiles are generated on a grid with overlap, and the outer regions of each prediction are discarded before stitching (also referred to as \textit{outer tiling}~\cite{Ashesh2022-ah}).
An alternative approach computes a weighted average of overlapping regions, with weights decreasing toward the border, as in nnU-Net~\cite{isensee2021nnu} or Cellpose~\cite{stringer2021cellpose}.

While averaging approaches are well-suited to segmentation, where predictions are smooth class-probability maps and minor blurring across seams is tolerable, they are not optimal for image restoration tasks, where they risk attenuating and blending the high-frequency structures the model is trained to recover.
Ashesh et al.~\cite{Ashesh2022-ah} proposed inner tiling (or \textit{inner padding}) as a refinement for the image-splitting setting: rather than using a larger patch size (\textit{outer padding}), which puts the model out of distribution with respect to its training patches when the RF is larger than the patches, predictions are made on patches of the training size, and only a central crop is kept for stitching. 

Finally, we note that dense sliding-window evaluation across shifted alignments is itself well-established in fully-convolutional prediction~\cite{Ciresan2012-oh, Sermanet2013-lw, long2015fully}.
However, it has never been applied to posterior models, where the averaging step performed for smooth stitching could leverage the posterior aggregation required by these models.

\paragraph{\textbf{Training-sided mitigation strategies.}} 
While the above strategies operate at inference time, a complementary line of work attempts to remove tiling artifacts at their source during training~\cite{Buglakova2025-no, Lahiani2019-qm}.
Both contributions in this category trace the artifact to per-tile feature normalization with instance normalization (InstanceNorm)~\cite{ulyanov2016instance}, whose statistics depend on the current tile and thus make the same pixel's representation tile-dependent, breaking seamless stitching irrespective of halo size.
Buglakova et al.~\cite{Buglakova2025-no} address this issue architecturally by replacing InstanceNorm with BatchRenorm to obtain globally accumulated statistics shared between training and inference, and report artifact-free predictions without loss of accuracy on segmentation tasks.
Lahiani et al.~\cite{Lahiani2019-qm} tackle the same root cause through the loss function, introducing a perceptual embedding consistency term that regularizes the generator against tile-dependent representations during tile-wise virtual staining with GANs.
Both require retraining and have so far been demonstrated only on deterministic models for the particular tasks of segmentation and virtual staining.
Therefore, they do not address the specific case of posterior models, where the same input can admit multiple plausible reconstructions, and sampling variability is thus a main source of stitching artifacts. 

\paragraph{\textbf{Quantifying stitching artifacts.}}
Most prior work evaluates stitching artifacts indirectly, through task metrics (\eg, PSNR, Dice) that report overall prediction quality but do not isolate the localized failure at tile seams~\cite{Reina2020-zz, Ashesh2022-ah}. 
Moreover, they require ground truth and a non-tiled baseline, which is clearly unavailable in settings where tiling is used.
Closer to the reference-free regime, Buglakova et al.~\cite{Buglakova2025-no} introduce \textit{tile mismatch} as the Dice score between the same region of different overlapping tiles.
Unfortunately, the method requires access to the tiles prior to stitching; \ie it doesn't directly apply to already-stitched images and, in the current formulation, is restricted to segmentation outputs. 
We instead target a reference-free, task-agnostic regime applicable to any tiled-inference pipeline without access to pre-stitching model outputs.
The closest conceptual analog is the no-reference JPEG-blockiness image quality assessment literature~\cite{Wang2002-ld, Wang2003-kk}, which compares pixel statistics at known block boundaries against those of surrounding interior regions~\cite{Pan2007-kq, Lee2012-ol}, with later refinements introducing perceptual masking~\cite{Liu2009-yz} and gradient-domain formulations~\cite{Perra2005-bh}.
Our construction adapts these ideas from fixed-size JPEG blocks to inference tiles of arbitrary geometry, replacing the heuristic ratio scoring with a non-parametric two-sample test~\cite{Chen2004-mm, Demidenko2004-cn} applied to local gradient distributions.

\section{Methods}
\label{sec:methods}

\begin{figure}[t!]
    \centering
    \includegraphics[width=\linewidth]{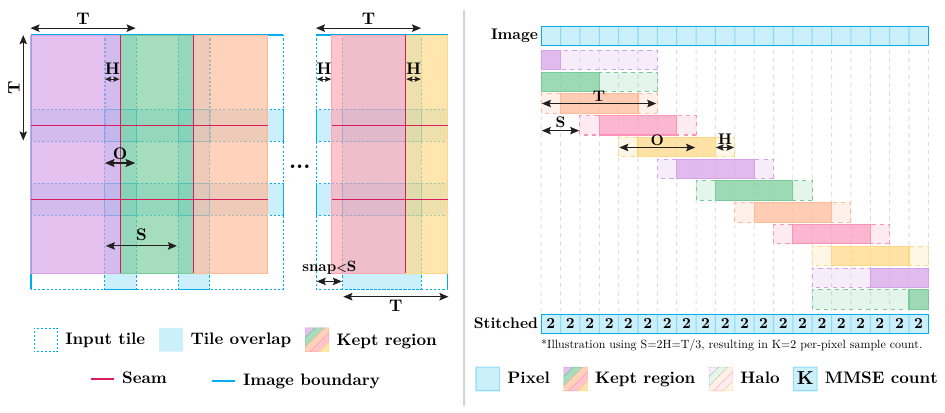}
    \caption{
        \textbf{Left: illustration of inner tiling in 2D. } 
        Each tile of size $T$ is fed to the model, and only its inner region, obtained by dropping halos of size $H$ on each side, is retained in the stitched output. 
        The halo of width $H$ (note: $H=O/2$ here, where $O$ is the physical overlap between tiles) is dropped from each side of each interior tile, so stitching seams are located exactly at the midpoint of the overlap zone. 
        Tiles are emitted at a fixed stride $S = T-2H$. 
        When the image extent is not a multiple of $S$, the last tile on each axis is snapped to the image edge, resulting in an effective stride less than $S$. Therefore the snapped tile contributes a narrower strip to the stitched output, which is from the previous seam to the image edge.
        \textbf{Right: illustration of SWITi along a single axis. }
        By introducing a stride $S < T - 2H$ while maintaining the same inner region geometry as in inner tiling, each pixel is covered by $K = \lfloor (T - 2H) / S \rfloor$ independently sampled tile predictions, which are averaged to get an MMSE estimate.
        \textit{Boundary handling:} We need to ensure that the sample count for all pixels is equal to $K$. 
        Therefore, at the image boundary thinner regions from the output tiles, clipped to the image extent, contribute to the sample average. 
        To avoid injecting synthetic pixels into the model, at the image boundaries, we select the input tiles to be those at the nearest valid coordinate; note that this means the contributing regions of these boundary tiles will not be the inner region.
    }
    \vspace{-2mm}
    \label{fig:1_tiling_strategies}
\end{figure}

We consider the problem of producing a dense prediction $\hat y \in \mathbb{R}^{C \times Z \times Y \times X}$ from an input image  $x$ of the same spatial extent, using a model $f_\theta$ whose receptive field and memory footprint preclude evaluating $f_\theta(x)$ in a single forward pass. 
The image is partitioned into overlapping spatial tiles; each tile is processed independently by $f_\theta$, and the per-tile predictions are recombined into a single output of the original size. 
For simplicity, we describe the tiling geometry along a single axis $D$; the construction can easily be extended to other axes by simply applying the exact same reasoning.

\subsection{Inner Tiling in Posterior Models}
\label{subsec:meth_inner_tiling}

\paragraph{\textbf{Inner Tiling Geometry.}}
Our implementation of the inner tiling approach is schematically illustrated in the left panel of~\cref{fig:1_tiling_strategies}.
We refer to the tile size along one spatial axis as $T$. 
Adjacent tiles overlap by $O$ pixels, so consecutive tile origins are separated by a stride of $S = T - O$.
In practice, from every predicted tile, we discard a halo of width $H=O/2$ on each side and retain only the \textit{inner region} of width $T-2H$.
Adjacent inner regions thus meet at the midpoint of the overlap zone, where the stitching seam lies. 
When the axis length $D$ is not a multiple of $S$, the last tile is anchored to the image edge, widening its overlap with the preceding tile (see~\cref{fig:1_tiling_strategies}).
Every output location is then written by exactly one tile, and $\hat y$ is the composition of all inner regions.

\paragraph{\textbf{MMSE predictions.}} 
For posterior models, each evaluation of $f_\theta$ draws a single sample from the predictive posterior $p(y \mid x)$ rather than a deterministic output, and the standard estimator at inference time is the empirical mean over $M$ stochastic forward passes per tile, \ie, the MMSE estimate. 
Increasing $M$ averages out the within-tile stochastic variability, but only at fixed tile positions: every repeated pass is computed on the same partition of the image, so the seams between inner regions remain at the same locations.

\sloppy

\paragraph{\textbf{Computational cost.}} 
Along each axis, the inner tiling strategy emits $\lceil D / (T - 2H) \rceil$ tiles. 
By collecting $M$ posterior samples for each tile of an $n$-dimensional image, the total number of forward passes through the model scales as $\mathcal{O}\!\left( \left(\tfrac{D}{T - 2H}\right)^{\!n} \cdot M \right)$.

\fussy

\subsection{Sliding-Window Inner Tiling (SWITi)}
\label{sec:switi}
SWITi breaks fixed-partition anchoring by sampling tiles at a stride shorter than the inner region width so that inner regions overlap.
Each pixel is then estimated by averaging the inner regions that cover it, drawn at different tile locations; seams are therefore redistributed across locations rather than confined to fixed coordinates and attenuated by construction (see~\cref{subsec:seam_amplitude_reduction}).

In practice, SWITi achieves this by promoting the stride $S$ to a free parameter while keeping the tile size $T$ and the halo $H$ (and thus the inner region width $T-2H$) unchanged~(\cref{fig:1_tiling_strategies}, right panel).
For $S \le T-2H$, the stride parameter controls the overlap between inner regions and thus the number of inner regions covering a given pixel, that is $K(S) = \lfloor (T-2H)/S \rfloor$ per-axis (hence $K(S)^n$ in $n$ spatial dimensions).
Therefore, for fixed $T$ and $H$, setting the stride along each axis to $S = \lfloor (T-2H)/K \rfloor = \lfloor (T-2H)/M^{1/n} \rfloor$,
allows matching the number $M$ of averaged samples at each location under MMSE estimation in the inner tiling.
In summary, both strategies average $M$ samples at the same forward-pass count, up to a sub-leading edge term, and only differ in where the samples are drawn: $M$ co-located draws become $M$ spatially shifted draws.

Two noteworthy points: $(i)$~for $S = T-2H$, we have $K(T-2H) = 1$: SWITi reduces to classical inner tiling, of which it is thus a strict generalization.
$(ii)$~By employing naive striding, pixels near the image edges would be covered by fewer than $K^n$ inner regions.
We avoid this by sampling additional tiles and clipping their inner regions to the image extent, which restores uniform coverage~(\cref{fig:1_tiling_strategies}).

\begin{figure}[t]
    \centering
    \includegraphics[width=\linewidth]{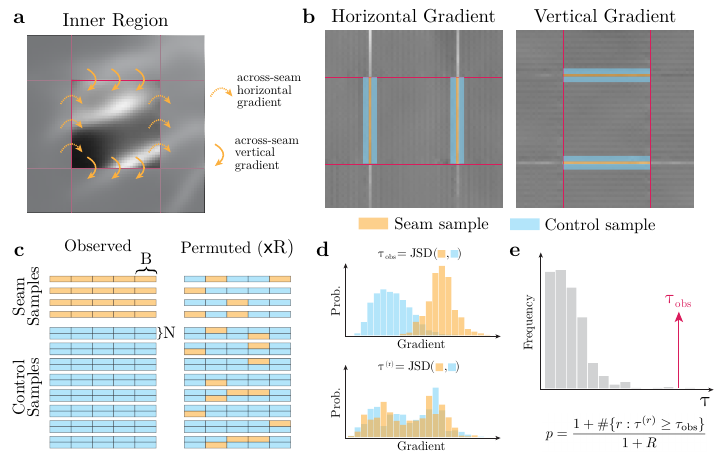}
    \caption{
        \textbf{Per-tile seam detection and quantification via permutation test.} 
        $(a)$~Within each retained inner region, we compute across-seam gradients in the horizontal and vertical directions. 
        $(b)$~At every seam, we draw a seam sample (on the seam) and a control sample (adjacent strips), separately for horizontal and vertical gradients. 
        $(c)$~Each sample is partitioned into blocks of size $B$; the seam/control labels are permuted $R$ times, preserving block counts. 
        $(d)$~The statistics $\tau_{\text{obs}}$ and $\tau^{(r)}$ are computed as the Jensen-Shannon Divergence between the distributions of seam and control samples, observed and permuted, respectively. 
        $(e)$~Comparing $\tau_{\text{obs}}$ against the permuted statistics yields the p-value.
    }
    \label{fig:2_permtest}
\end{figure}

\subsection{FRT \& ASV: artifact metrics from per-tile permutation tests}
\label{sec:meth_metric}

We accompany SWITi with a method to detect and quantify stitching artifacts directly from a stitched predicted image, without the need for ground truth.
For the inner region of each tile, we perform a two-sample statistical test comparing the distributions of adjacent-pixel gradients taken \emph{across the seams} with those of the immediately neighboring pixels.
Under the null hypothesis of artifact-free stitching, the two distributions are statistically indistinguishable.
Conversely, a significant difference between the distributions signals a discontinuity at the seam that the underlying image content does not account for.
For each retained inner region, we compute the $p$-value and the normalized test statistic of the observed data.
This enables simultaneous detection, quantification, and spatial localization of artifacts.
The fundamental aspects of the proposed method are illustrated in~\cref{fig:2_permtest}.

\paragraph{\textbf{Seam and control samples.}}
An across-seam gradient is the one-step finite difference between two adjacent pixels on either side of a seam, that is, the change in pixel value in the direction perpendicular to the seam~(\cref{fig:2_permtest}a).
The \emph{seam sample} of an inner region pools the directional across-seam gradients over every position along each of its (up to) four seams (at the region's edges).
The gradient \emph{control sample} is drawn from the same local neighborhood, taking parallel strips of width $N$ on each side of the seam~(\cref{fig:2_permtest}b).
In this way, the control shares a similar image content as the seam, ensuring any difference between the two distributions reflects the stitching discontinuity rather than a mismatch in underlying image content or texture.
Moreover, we sample both sides of the boundary to prevent the control from being biased toward either of the two independently predicted inner regions.
The test applies unchanged in 3D: in that case, the seam samples are collected along the (up to) six two-dimensional faces that define boundaries between neighboring inner regions.

All directional gradients are pooled into a single sample.
For anisotropic data (\eg coarser axial than lateral sampling in 3D), we first standardize gradients per axis and balance each axis's contribution so that no direction dominates; see~\cref{sec:suppl_anisotropy} for details.

\paragraph{\textbf{Block permutation test.}}
We assess the discrepancy between the seam and control samples using a permutation test, which empirically builds the null distribution of the test statistic by resampling the pooled data.
This gives us flexibility in the choice of the test statistic: we can pick one that is more sensitive to the seam discontinuity without restricting ourselves to statistics with a tractable null distribution.
Permuting individual pixel gradients, however, would assume they are exchangeable under the null hypothesis.
This assumption does not hold because gradients from nearby pixels are spatially correlated through shared image content; treating them as exchangeable would make the test anti-conservative (\ie; it would reject the null too often, see~\cref{sec:calibration}).
We therefore permute contiguous blocks of $B$ gradients rather than individual values, preserving the local correlation structure within each block and exchanging only the block labels~(\cref{fig:2_permtest}c).

Concretely, the block permutation test is performed as follows: first, we compute the observed statistic $\tau_{\text{obs}}$ on the original distributions.
Then, the blocks in the seam and control samples are permuted between the two groups $R$ times, with each permutation selected uniformly at random. 
For each permutation ($r=1,2,\ldots,R$) we compute the corresponding statistic $\tau^{(r)}$(\cref{fig:2_permtest}c,d). 
The empirical $p$-value is the fraction of permuted statistics that are at least as large as the observed one~(\cref{fig:2_permtest}e), \ie, 
\begin{equation}
    \label{eq:pvalue}
    p = \frac{1 + \#\{r : \tau^{(r)} \geq \tau_{\text{obs}}\}}{1 + R}.
\end{equation}
The $+1$ offsets implement the Phipson and Smyth correction~\cite{Phipson2010-rw}, which guarantees a valid, strictly positive $p$-value.
As the default statistic, we use the Jensen--Shannon divergence between the empirical histograms of the seam and control samples, chosen for its symmetry and boundedness.
The Kullback--Leibler divergence, the Kolmogorov--Smirnov statistic, and the Wasserstein-1 distance are reported as ablation in the supplement~(\cref{subsec:statistic_ablation}). 

\paragraph{\textbf{Aggregation and reporting.}}
Each stitched predicted image yields a list of per-inner-tile $(\tau_{\text{obs}}, p)$ values.
In the results, we instead report the normalized test statistic $Z_{\text{obs}}$ for a given inner tile, which is obtained as the $Z$-score of $\tau_{\text{obs}}$ with respect to the sample mean and standard deviation of the null statistic $\tau^{(r)}$.
This ensures that the test statistic's value is comparable across tiles and images.
Therefore, we can derive: $(i)$~an artifact \textit{detection} metric, which is the per-image fraction of tiles with $p < \alpha$ at a significance level $\alpha$: we call this the \textbf{F}raction of \textbf{R}ejected \textbf{T}ests (FRT); $(ii)$~a metric that quantifies artifact \textit{severity}, which is the median value of $Z_{\text{obs}}$ across an image: the higher $Z_{\text{obs}}$, the greater the dissimilarity of the seam and control samples and, hence, the severity of the seam.
We call this metric \textbf{A}rtifact \textbf{S}e\textbf{V}erity (ASV).

Finally, we flag a failure mode of this method: a model that produces uniformly over-smoothed predictions collapses both the seam and the control gradient distributions toward zero and will therefore achieve low $\tau_{\text{obs}}$ everywhere, indistinguishably from artifact-free stitching.
The metric needs to be reported jointly with a standard reconstruction-quality measure so that improvements in one cannot be obtained at the expense of the other.
\section{Experimental Setup}
\label{sec:exp_setup}

\begin{table*}[t!]
\centering
\caption{Per-tile stitching-artifact statistics for inner tiling and SWITi across datasets,
at control-strip width $N=2$.
FRT is the fraction of tiles whose per-tile permutation $p$-value falls below $\alpha=0.05$ across one image; ASV is the median permutation
$z$-score of the permutation test statistic ($\mathrm{median}(Z_{\mathrm{obs})})$.
Values are averaged over test images; $\pm$ denotes standard error across images (except for CBG-Z18, whose test set comprises a single volume).
Lower values indicate fewer / less severe detected tiling artifacts.}
\label{tab:permutation_results}
\setlength{\tabcolsep}{4pt}
\begin{tabular}{lll cc}
\toprule
Dataset & Method & Channel & FRT \ $\downarrow$ &
ASV $\downarrow$ \\
\midrule
\multirow{4}{*}{PaviaATN}
  & \multirow{2}{*}{Inner Tiling} & Ch.\,1 & $0.965 \pm 0.002$ & $14.90 \pm 0.16$ \\
  &                               & Ch.\,2 & $0.993 \pm 0.000$ & $26.17 \pm 0.12$ \\
\cmidrule(l){2-5}
  & \multirow{2}{*}{SWITi}        & Ch.\,1 & $\mathbf{0.497 \pm 0.007}$ & $\mathbf{1.78 \pm 0.06}$ \\
  &                               & Ch.\,2 & $\mathbf{0.771 \pm 0.003}$ & $\mathbf{5.41 \pm 0.11}$ \\
\midrule
\multirow{6}{*}{CBG-Z18}
  & \multirow{3}{*}{Inner Tiling} & Ch.\,1 & $0.877$ & $12.26$ \\
  &                               & Ch.\,2 & $0.572$ & $2.56$ \\
  &                               & Ch.\,3 & $0.775$ & $7.28$ \\
\cmidrule(l){2-5}
  & \multirow{3}{*}{SWITi}        & Ch.\,1 & $\mathbf{0.468}$ & $\mathbf{1.53}$ \\
  &                               & Ch.\,2 & $\mathbf{0.265}$ & $\mathbf{0.33}$ \\
  &                               & Ch.\,3 & $\mathbf{0.431}$ & $\mathbf{1.23}$ \\
\midrule
\multirow{6}{*}{HT-LIF24 (5\,ms)}
  & \multirow{3}{*}{Inner Tiling} & Ch.\,1 & $0.816 \pm 0.008$ & $5.44 \pm 0.23$ \\
  &                               & Ch.\,2 & $0.811 \pm 0.014$ & $\mathbf{5.27 \pm 0.28}$ \\
  &                               & Ch.\,3 & $0.815 \pm 0.010$ & $5.92 \pm 0.26$ \\
\cmidrule(l){2-5}
  & \multirow{3}{*}{SWITi}        & Ch.\,1 & $\mathbf{0.479 \pm 0.030}$ & $\mathbf{1.55 \pm 0.34}$ \\
  &                               & Ch.\,2 & $\mathbf{0.686 \pm 0.021}$ & $5.79 \pm 0.70$ \\
  &                               & Ch.\,3 & $\mathbf{0.520 \pm 0.023}$ & $\mathbf{1.96 \pm 0.16}$ \\
\bottomrule
\end{tabular}
\end{table*}

\begin{table}[t]
\centering
\caption{Resolution via Fourier Ring Correlation (FRC) at the fixed $1/7$ threshold,
for inner tiling and SWITi.
We report the cutoff spatial frequency (cyc/px; higher is better) as the per-channel
mean $\pm$ standard error over images.
Per row, the better method is shown in bold.}
\label{tab:frc}
\setlength{\tabcolsep}{4pt}
\begin{tabular}{llcc}
\toprule
 & & \multicolumn{2}{c}{Sp.\ freq.\ @ $\text{FRC}=1/7$ (cyc/px) $\uparrow$} \\
\cmidrule(lr){3-4}
Dataset & Channel & Inner Tiling & SWITi \\
\midrule
\multirow{3}{*}{PaviaATN}
 & Ch. 1  & $0.128 \pm 0.002$ & $\mathbf{0.151 \pm 0.002}$ \\
 & Ch. 2  & $0.025 \pm 0.002$ & $\mathbf{0.049 \pm 0.008}$ \\
\midrule
\multirow{4}{*}{CBG-Z18}
 & Ch. 1  & $0.264 \pm 0.005$ & $\mathbf{0.273 \pm 0.004}$ \\
 & Ch. 2  & $0.305 \pm 0.004$ & $\mathbf{0.313 \pm 0.004}$ \\
 & Ch. 3  & $0.311 \pm 0.010$ & $\mathbf{0.329 \pm 0.007}$ \\
\midrule
\multirow{4}{*}{HT-LIF24 (5 ms)}
 & Ch. 1  & $0.150 \pm 0.003$ & $\mathbf{0.160 \pm 0.004}$ \\
 & Ch. 2  & $0.233 \pm 0.006$ & $\mathbf{0.234 \pm 0.006}$ \\
 & Ch. 3  & $0.273 \pm 0.009$ & $\mathbf{0.276 \pm 0.010}$ \\
\bottomrule
\end{tabular}
\end{table}
\begin{figure}[bth]
    \centering
    \includegraphics[width=\linewidth]{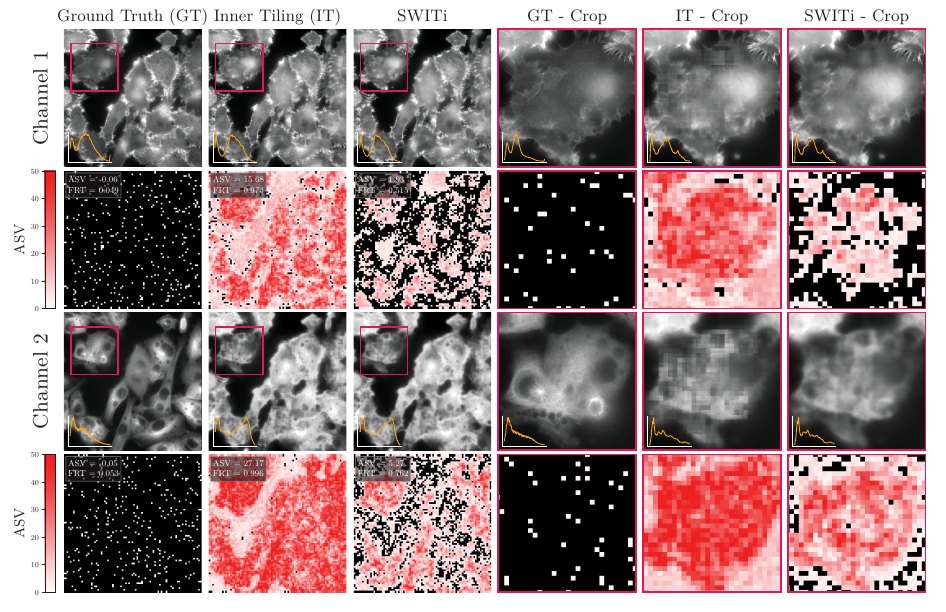}
    \caption{
        \textbf{Qualitative comparison and per-tile seam detection on PaviaATN},
        for both split channels.
        Per channel: top row, predictions (inset: intensity histogram); bottom row, per-tile permutation-test significance maps, each tile colored by its ASV score, with tiles above the significance threshold ($p \ge \alpha$, $\alpha = 0.05$) left uncolored (inset box: per-image test results).
        Magenta boxes mark the crop regions.
        Inner tiling shows clearly visible, grid-aligned seams, whereas SWITi prediction exhibits a coherent, seam-free texture.
        The permutation test flags the inner tiling seams as prevalent and severe, and assigns SWITi markedly lower values at the same coordinates, confirming that the inner tiling artifacts are correctly detected and measured.
    }
    \label{fig:3_quali_Pavia}
\end{figure}

We evaluate SWITi on image splitting of fluorescence microscopy data using Micro$\mathbb{S}$plit~\cite{Ashesh2026-da}, the state-of-the-art model.
Its hierarchical-VAE architecture makes it particularly prone to stitching artifacts for the following reasons: first, as a posterior model, it can sample mutually inconsistent reconstructions across neighboring tiles. 
Second, its hierarchical depth, combined with lateral-context inputs, gives the model an effective receptive field far larger than the tile size that would fit in a common GPU. 
As a result, traditional inner tiling cannot remove tiling artifacts on its own. 
SWITi, on the other hand, is specifically designed to address and mitigate these shortcomings.
In all the experiments that follow, we use pre-trained, publicly available checkpoints from Micro$\mathbb{S}$plit.
MMSE estimates are obtained by averaging $M=64$ posterior samples for both inner tiling and SWITi.

\paragraph{\bf Datasets. }
We evaluate SWITi on Micro$\mathbb{S}$plit benchmarks that exhibit visible tiling artifacts under inner tiling, across a range of settings: spatial dimensionality (2D and 3D), input formation (synthetic vs. directly acquired superimposed images), number of unmixed structures, and amount of noise.
The datasets included are the following:
\noindent\textbf{PaviaATN}~\cite{Ashesh2022-ah} is 2D fluorescence microscopy of cells labeling actin and tubulin, with superimposed inputs synthesized by channel summation.
\textbf{CBG-Z18}~\cite{Weigert2018-jr} provides 3D spinning-disk confocal volumes of zebrafish retinal tissue; the three-structure task separates nuclei, nuclear envelope, and cell membrane from synthetically summed inputs.
\textbf{HT-LIF24}~\cite{Ashesh2026-da} is a 2D spinning-disk confocal of cultured cells (microtubules, nuclei, kinetochores), with superimposed inputs acquired directly at the microscope; of its several exposures, we use the highest-noise 5\,ms regime.

\paragraph{\bf Evaluation Metrics. }
To detect, localize, and quantify tiling artifacts, we use our reference-free per-tile permutation test, as described in~\ref{sec:meth_metric}.
We use the Jensen--Shannon divergence as the test statistic, a block size $B=3$ to ensure the calibration of the test, a strip width of $N=2$ for the control samples, and $R=1000$ permutations.
Since the test is run for each inner region, detection and localization are achieved by visualizing the positions of regions for which the test rejected the null hypothesis.
Moreover, we use the Fraction of Rejected Tests (FRT) and the Artifact Severity (ASV) scores to quantify, respectively, the number and severity of tiling artifacts. 

As previously mentioned, the test can fail due to over-smoothing, which could remove seams along with detail.
For this reason, we report it alongside global metrics that certify overall reconstruction quality: Range-Invariant PSNR~\cite{Weigert2018-jr}, MS-SSIM~\cite{wang2003msssim} and Pearson's Correlation Coefficient for fidelity, and LPIPS~\cite{Zhang2018-xx} for perceptual quality. 

At a matched computational budget and retained area per tile, the permutation test lets us compare inner tiling and SWITi at inner tiling's seam coordinates. 
This comparison is informative but coordinate-bound: it cannot, on its own, rule out that artifacts are redistributed rather than removed. 
We therefore adopt a location-agnostic comparison based on Fourier Ring Correlation (FRC)~\cite{Saxton1982-wy, van-Heel2005-nl}.
FRC measures the agreement between the spatial frequencies of two images across rings and is common in microscopy.
Here, it is computed between the predicted and ground-truth images.
In this context, FRC allows to detect and measure the spurious high-frequency content introduced by tiled predictions.
We summarize FRC curves by the spatial frequency at which it crosses the fixed $1/7$ threshold, which is commonly used in literature as a resolution estimate~\cite{Rosenthal2003-ia, Nieuwenhuizen2013-lp, Koho2019-wn}.
In 3D, we compute FRC on individual $xy$ slices rather than the volumetric Fourier Shell Correlation, whose shells average over directions of differing effective resolution under anisotropic sampling~\cite{Koho2019-wn}.
\section{Results}
\label{sec:results}

\begin{figure}[bt]
    \centering
    \includegraphics[width=\linewidth]{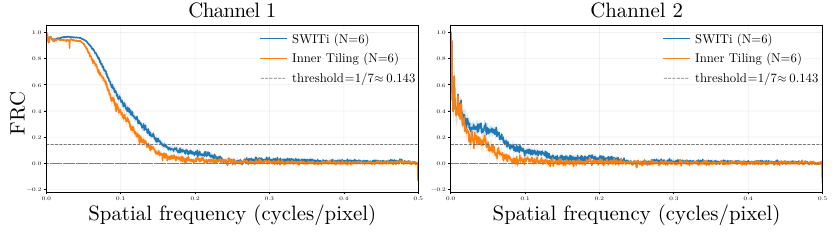}
    \caption{
        \textbf{Fourier Ring Correlation (FRC) curve on PaviaATN.}
        FRC between the prediction and the ground truth (GT) reference as a function of spatial frequency, for the two split channels, comparing inner tiling (orange) and SWITi (blue); shaded bands denote the standard error over $N=6$ images.
        The dashed line marks the fixed $1/7 \approx 0.143$ threshold, whose crossing defines the resolution cutoff (the higher, the better).
        In both channels, the SWITi curve lies at or above inner tiling across the informative frequency band and crosses the threshold at a higher frequency, indicating that SWITi remains more faithful to GT also at higher spatial frequencies.
        Since FRC is agnostic to the location of stitching seams, these results confirm that SWITi delivers artifact reduction and not redistribution.
    }
    \label{fig:4_FRC_Pavia}
\end{figure}

We compare SWITi against MMSE-based inner tiling across three splitting benchmarks, evaluating both the qualitative appearance of stitched predictions and the quantitative metrics described in~\cref{sec:exp_setup}.

\paragraph{\textbf{Qualitative comparison.}}
In~\cref{fig:3_quali_Pavia}, \ref{fig:suppl_quali_CBGZ18}, and \ref{fig:suppl_quali_HT5ms} we visually compare inner tiling and SWITi predictions (first row for each channel).
Inner tiling produces conspicuous, grid-aligned artifacts, particularly evident in the insets.
These discontinuities make tiled posterior predictions difficult to read and to analyze downstream, as a seam can be mistaken for an actual boundary between biological structures.
Under SWITi, the same regions are recovered more coherently, and the recombined image is shown as a single homogeneous acquisition rather than a mosaic of tiles.

\paragraph{\textbf{Per-tile seam detection and quantification.}}
This visual impression is quantitatively corroborated by the per-tile gradient permutation test results~(\cref{tab:permutation_results} and ~\cref{fig:3_quali_Pavia}, \ref{fig:suppl_quali_CBGZ18}, and \ref{fig:suppl_quali_HT5ms}, second row for each channel).
Results show that SWITi consistently reduces the presence of stitching artifacts in every dataset, both in terms of prevalence (lower fraction of rejected tests, FRT, for each image) and severity (lower ASV score).
In turn, the results also validate the proposed metrics: indeed, both FRT and ASV score highest for the PaviaATN dataset, which exhibits more marked tiling artifacts.

Because the test is evaluated on the inner tiling grid, the better metrics for SWITi are direct evidence of attenuation at those coordinates, but they cannot, on their own, separate removal from redistribution, nor exclude over-smoothing (\cref{sec:meth_metric}).
We therefore need to read them together with global fidelity and a location-agnostic resolution measure.

\paragraph{\textbf{Global reconstruction quality.}}
Global metrics~(\cref{tab:global_results}) are generally comparable between the two methods across every dataset.
The near-parity in metrics is expected: stitching seams are spatially localized, so they contribute little to pixel-aggregate scores regardless of whether they are present; this is exactly why such metrics cannot serve as primary evidence for seam removal.
More precisely, SWITi shows slightly better performance on fidelity metrics (PSNR, MS-SSIM, Pearson correlation coefficient), while inner tiling is marginally better on LPIPS. 
The reason for this is plausibly attributable to the perception–distortion tradeoff~\cite{Blau2018-cj}, as the denser posterior averaging in SWITi trades a small amount of perceptual quality for closer agreement with the reference.
Critically, these results rule out the failure mode of the per-tile test.

\paragraph{\textbf{Resolution comparison via FRC.}}
To compare the two methods without conditioning on the inner tiling seam grid, we employ the Fourier Ring Correlation~(\cref{tab:frc}, \cref{fig:4_FRC_Pavia} and~\ref{fig:suppl_FRC_others}).
The cutoff spatial frequency at the $1/7$ threshold is consistently higher for SWITi than for inner tiling, with the clearest gains on the datasets that exhibit the most pronounced seams, PaviaATN and CBG-Z18.
Since seam discontinuities inject spurious high-frequency content that does not correlate with the reference, the observed increase in FRC confirms that SWITi significantly mitigates their impact.

\paragraph{}
Taken together, all the presented results point to a consistent conclusion: at matched inference cost, SWITi substantially attenuates stitching seams while preserving, and in several cases improving, reconstruction fidelity and resolution.
\section{Discussion}
\label{sec:discussion}

Tiled inference produces artifacts from two sources: a deterministic one from partial receptive field content and a stochastic one from independent posterior sampling.
While a suitably large halo can remove artifacts for deterministic networks with a moderately sized receptive field, no method exists that can fully eliminate either of these two sources of tiling artifacts for posterior models with a large RF.
Enlarging the computational budget and averaging more posterior samples into an increasingly better MMSE prediction only reduces the stochastic part, which, for the PaviaATN data, accounts for around $7.5\%$ of the residual inner tiling seam, as can also be seen in~\cref{subsec:seams_deterministic}.

SWITi changes the spatial context from which individual samples are inferred.
In this way, inner regions from gradually shifted tiles are combined such that a discontinuity across a seam, both deterministic and stochastic, is attenuated by the per-pixel coverage $K$, as shown in~\cref{subsec:seam_amplitude_reduction}.
This can be achieved with the same forward-pass count as standard MMSE inner tiling and therefore does not incur additional computation\footnote{Please note that the cost-neutrality of SWITi is specific to posterior models.
Applied to a deterministic network, sliding-window averaging has no posterior draws to reallocate and collapses to the averaging of overlapping predictions.}.

The per-tile permutation test we introduced detects, localizes, and quantifies tiling artifacts along tile seams from the stitched prediction alone, needing neither ground truth nor the individual pre-stitching tiles, nor any re-training or re-inference steps. From its per-tile outcomes, we derive two reference-free metrics: the Fraction of Rejected Tests (FRT), which reports how prevalent seams are, and Artifact Severity (ASV), which reports how severe they are. We showed that the test is well calibrated on seam-free references (see~\cref{sec:calibration}), and the severity ASV measures, in all our experiments, align with the visual saliency of tiling artifacts, as shown in~\cref{fig:3_quali_Pavia,tab:permutation_results}.
Our test cannot, by itself, rule out a redistribution of artifacts, nor can it exclude the possibility of being satisfied by a uniformly over-smoothed prediction that flattens all gradients, from artifacts or true structure alike.
We therefore cross-checked using two complementary measures: global fidelity (PSNR, MS-SSIM, Pearson), which would degrade if detail were lost to over-smoothing, and Fourier Ring Correlation (FRC), which is agnostic to the location of tile seams.
Fidelity remains essentially unchanged between the two methods, while the FRC cutoff improves~(\cref{tab:frc,fig:4_FRC_Pavia}), showing that SWITi consistently attenuates tiling artifacts.

Whenever posterior models are applied to images that are too large to process in one pass, SWITi and its companion test can be applied at no additional cost and without retraining.
Beyond image splitting, the test requires only a stitched prediction and thus applies equally to any tiled inference pipeline; we hope it will be adopted as a general diagnostic for stitching artifacts.
Whether a comparably cost-neutral redistribution can be found for deterministic models, where sliding-window averaging currently adds forward passes, remains open.
An implementation of our work will be released publicly under a permissive open source license, allowing everyone to benefit from SWITi and the reference-free tiling artifact detection and quantification methods we have introduced here.

\section*{Acknowledgements}
This work was supported by the European Union through the Horizon Europe program (IMAGINE project, grant agreement 101094250-IMAGINE).
Additionally, the authors want to thank all members of the Image Analysis Facility and the Jug Group at Human Technopole for useful feedback and discussions and the IT and HPC teams at HT for the compute infrastructure they make available to us.


%
%
\bibliographystyle{splncs04}
\bibliography{references}

\clearpage

\newpage
\begin{center}
\Large
\textbf{\textit{Supplementary Material}} \\
\vspace{4mm}
\textbf{
SWITi: Quantifying and Reducing Tiling Artifacts
with Sliding Window Inner Tiling
}\\[.2cm]
\end{center}
\appendix

\setcounter{page}{1}
\setcounter{section}{0}
\setcounter{figure}{0}
\setcounter{table}{0}
\setcounter{equation}{0}
\renewcommand{\thepage}{S.\arabic{page}} 
\renewcommand*{\thesection}{S.\arabic{section}}
\renewcommand{\thefigure}{S.\arabic{figure}}
\renewcommand{\thetable}{S.\arabic{table}}
\renewcommand{\theequation}{S.\arabic{equation}}

\section{Extended Methods}

\subsection{Handling Anisotropic Gradient Statistics}
\label{sec:suppl_anisotropy}

The per-tile test pools directional across-seam gradients into a single seam sample and a single control sample.
This pooling is valid only if the directional gradients share a common distribution under the null; on anisotropic data, two asymmetries break that assumption, and we correct both once, before any permutation, so that seam and control samples remain exchangeable.

\paragraph{Per-axis magnitude.}
Microscopy data are often anisotropic: in 3D, for instance, axial sampling is typically coarser than lateral sampling, so gradients computed along $z$ are systematically larger in magnitude than those computed along $x$ or $y$.
Pooling directional gradients across axes without correction would let the axis with the largest gradients dominate the test statistic.
We therefore standardize directional gradients per axis, using the mean and standard deviation computed over the whole image, so that gradients from different axes lie on a common scale and can be pooled into a single sample.

\paragraph{Per-axis count.}
A second asymmetry arises when the tile size differs across dimensions: the extent of the seam strips (in 2D) or faces (in 3D) then differs across axes, and some axes contribute far more gradients than others.
We subsample the over-represented axes so that all directions are equally weighted in the seam sample.

\subsection{Tiling Artifact Attenuation with SWITi}
\label{subsec:seam_amplitude_reduction}

The aim of SWITi is to reduce the amplitude of the artifact that occurs at a tile boundary due to inter-tile differences. 
In this section, we show that, compared to inner tiling, SWITi reduces the contribution of inter-tile differences by a factor of $K$, where $K$ is the number of tiles overlapping a pixel. \newline

We will consider an input image $X$, made up of pixels $x_i \in X$, where $i \in I$ is an index. We will denote the prediction of the model for a particular pixel as $y_j(x_i)$ where each $j \in  J$ denotes a particular output tile. 
Each output tile $j$ is produced from a different input tile location in $X$ and by a separate forward pass through the model; therefore, inter-tile differences arise from both previously discussed sources: one being the finite receptive field of CNNs and the second being the stochastic nature of sampling from a posterior distribution. 
Note that each $y_j$ is only defined for a subset of pixels in $X$.

\paragraph{Adjacent-pixel difference.} Consider the adjacent pixels $x_i$ and $x_{i+1}$, then let $\varepsilon$ be the smallest value for which the following holds:
\[
  |y_j(x_{i+1}) - y_j(x_i)| \leq \varepsilon \qquad \textrm{$\forall j$ such that $y_j$ is defined for both $x_i$ and $x_{i+1}$},
\]
noting that the L.H.S. is the magnitude of the difference between adjacent pixels in the \emph{same} output tile. We can see that $\varepsilon$ is  the largest such difference between adjacent pixels. Following from the assumption that the outputs of a CNN are ``smooth``, $\varepsilon$ will always be relatively small. 

\paragraph{Seam attenuation under sliding-window averaging.} The set $J_i \subseteq J$ denotes $J_i = \{j :  \textrm{$y_j$ is defined for $x_i$\}}$, that is, $J_i$ defines the set of tiles that cover $x_i$. Let the size of the set $J_i$ be $|J_i| = K$; SWITi is formulated so that this is true for all $i \in I$. Now we can write the SWITi output for the pixel $x_i$ as
\[
y(x_i) = \frac{1}{K} \sum_{j \in J_i} y_j(x_i),
\]
so, it follows that the magnitude of the difference between two adjacent pixels is:
\[
|\mathrm{\Delta} y| = |y(x_{i+1}) - y(x_i)| = \frac{1}{K} \Big| \sum_{j \in J_{i+1}} y_j(x_{i+1}) - \sum_{j \in J_i} y_j(x_i) \Big|.
\]

Two cases arise, depending on whether an inner region boundary falls between $x_i$ and $x_{i+1}$.
If no boundary intervenes, the covering set is unchanged so, $J_i = J_{i+1}$. If a boundary does intervene, exactly one region is exchanged: a shared set $J_c = J_i \cap J_{i+1} $ of $K-1$ tiles covers both pixels, while tile $d$ (dropped) covers only $x_i$ and tile $a$ (added) covers only $x_{i+1}$, so $J_i = J_c \cup \{d\}$ and $J_{i+1} = J_c \cup \{a\}$.

\paragraph{Case 1: No boundary intervenes so $J_i = J_{i+1}$.} We can see that
\[
|\mathrm{\Delta} y | = \frac{1}{K} \Big| \sum_{j \in J_i} \big( y_j(x_{i+1}) - y_j(x_i) \big) \Big|.
\]
By the triangle inequality (TI), it follows that
\[
     |\mathrm{\Delta} y | \leq \frac{1}{K} \sum_{j \in J_i} \big| y_j(x_{i+1}) - y_j(x_i) \big|
     \leq \varepsilon.
\]
So we can conclude that if the magnitudes of all intra-tile differences are less than $\varepsilon$, then their averages are also less than $\varepsilon$; which is to say if the output of a CNN is smooth then an average of multiple outputs is also smooth.

\paragraph{Case 2: A Boundary intervenes so $J_i = J_c \cup \{d\}$ and $J_{i+1} = J_c \cup \{a\}$.} We will differentiate this output from the previous case by denoting it as $\mathrm{\Delta} y_\textrm{seam}$, we see that
\begin{align}
    |\mathrm{\Delta} y_\textrm{seam}| &= \Big| \frac{1}{K} \sum_{j \in J_c} \big[ y_j(x_{i+1}) - y_j(x_i) \big] + \frac{1}{K} \big[ y_a(x_{i+1}) - y_d(x_i) \big] \Big| \notag\\
    \shortintertext{by the TI,}
    &\leq \frac{1}{K} \sum_{j \in J_c} \big| y_j(x_{i+1}) - y_j(x_i) \big| + \frac{1}{K} \big| y_a(x_{i+1}) - y_d(x_i)  \big| \notag\\
    &\leq \frac{K - 1}{K} \varepsilon + \frac{1}{K} \big| y_a(x_{i+1}) - y_d(x_i)  \big|. \notag\\
    \shortintertext{and to simplify the expression,}
     &\leq \varepsilon + \frac{1}{K} \big| y_a(x_{i+1}) - y_d(x_i)  \big|. \label{eq:seam_switi}
\end{align}
The outputs $y_a(x_{i+1})$ and $y_d(x_i)$ come from two different tiles, so no assumptions can be made about the magnitude of their difference; however, we see that its contribution to $|\mathrm{\Delta} y_\textrm{seam}|$ is scaled by a factor of $\frac{1}{K}$.

\paragraph{Comparison with inner tiling.} Finally, we can compare the output of adjacent pixels over a seam boundary for SWITi and classical inner tiling. Classical inner tiling (without the MMSE estimate) is the case where $K = 1$ so $J_c = \emptyset$. We will consider the partition of tiles that include $a$ and $d$, which directly abut; therefore,
\begin{equation}\label{eq:seam_innertiling}
    |\mathrm{\Delta} y_\textrm{seam}^\textrm{(inner tiling)}| = |y_a(x_{i+1}) - y_d(x_i)|
\end{equation}
is a seam of full amplitude. We can substitute this expression \ref{eq:seam_innertiling} into \ref{eq:seam_switi}, yielding
\[
    |\mathrm{\Delta} y_\textrm{seam}^\textrm{(SWITi)}| \leq \varepsilon + \frac{1}{K}|\mathrm{\Delta} y_\textrm{seam}^\textrm{(inner tiling)}|, \notag\\
\]
where $K$ is still the number of tiles that overlap in SWITi. If there is a visible seam discontinuity for classical inner tiling, we can assume that $\varepsilon \ll |\mathrm{\Delta} y_\textrm{seam}^\textrm{(inner tiling)}|$, and therefore we can conclude that SWITi reduces the seam amplitude by a factor of $K$.

\section{Additional Results}
\label{sec:additional_results}

\begin{table*}[t!]
\centering
\caption{Global reconstruction metrics for inner tiling and SWITi across datasets.
Per-channel and average scores are reported for each method.
Arrows denote whether higher ($\uparrow$) or lower ($\downarrow$) is better;
per row, the better method is shown in \textbf{bold}.}
\label{tab:global_results}
\setlength{\tabcolsep}{4pt}
\begin{tabular}{ll c cccc}
\toprule
Dataset & Method & Channel & PSNR $\uparrow$ & MS-SSIM $\uparrow$ & LPIPS $\downarrow$ & Pearson $\uparrow$ \\
\midrule
\multirow{6}{*}{PaviaATN}
  & \multirow{3}{*}{Inner Tiling}
        & Ch.\,1 & 23.00 & 0.801 & \textbf{0.755} & 0.936 \\
  &     & Ch.\,2 & 20.64 & 0.689 & \textbf{0.714} & 0.691 \\
  &     & Avg.   & 21.82 & 0.745 & \textbf{0.735} & 0.814 \\
\cmidrule(l){2-7}
  & \multirow{3}{*}{SWITi}
        & Ch.\,1 & \textbf{23.11} & \textbf{0.817} & 0.821 & \textbf{0.938} \\
  &     & Ch.\,2 & \textbf{20.67} & \textbf{0.697} & 0.727 & \textbf{0.693} \\
  &     & Avg.   & \textbf{21.89} & \textbf{0.757} & 0.774 & \textbf{0.816} \\
\midrule
\multirow{8}{*}{CBG-Z18}
  & \multirow{4}{*}{Inner Tiling}
        & Ch.\,1 & 28.56 & 0.936 & \textbf{0.563} & 0.975 \\
  &     & Ch.\,2 & 28.95 & 0.922 & \textbf{0.349} & 0.945 \\
  &     & Ch.\,3 & 28.61 & 0.899 & 0.402 & 0.945 \\
  &     & Avg.   & 28.71 & 0.919 & \textbf{0.438} & 0.955 \\
\cmidrule(l){2-7}
  & \multirow{4}{*}{SWITi}
        & Ch.\,1 & \textbf{28.87} & \textbf{0.943} & 0.567 & \textbf{0.977} \\
  &     & Ch.\,2 & \textbf{29.33} & \textbf{0.927} & 0.354 & \textbf{0.950} \\
  &     & Ch.\,3 & \textbf{28.92} & \textbf{0.908} & \textbf{0.401} & \textbf{0.949} \\
  &     & Avg.   & \textbf{29.04} & \textbf{0.926} & 0.440 & \textbf{0.959} \\
\midrule
\multirow{8}{*}{HT-LIF24 (5\,ms)}
  & \multirow{4}{*}{Inner Tiling}
        & Ch.\,1 & 32.90 & 0.954 & \textbf{0.054} & 0.938 \\
  &     & Ch.\,2 & 34.33 & 0.982 & \textbf{0.014} & 0.948 \\
  &     & Ch.\,3 & 37.59 & 0.982 & \textbf{0.027} & 0.894 \\
  &     & Avg.   & 34.94 & 0.973 & \textbf{0.032} & 0.926 \\
\cmidrule(l){2-7}
  & \multirow{4}{*}{SWITi}
        & Ch.\,1 & \textbf{32.98} & \textbf{0.955} & 0.059 & \textbf{0.939} \\
  &     & Ch.\,2 & \textbf{34.38} & 0.982 & 0.015 & \textbf{0.949} \\
  &     & Ch.\,3 & \textbf{37.63} & 0.982 & 0.028 & 0.894 \\
  &     & Avg.   & \textbf{34.99} & 0.973 & 0.034 & \textbf{0.927} \\
\bottomrule
\end{tabular}
\end{table*}

\begin{figure}[t]
    \centering
    \includegraphics[width=\linewidth]{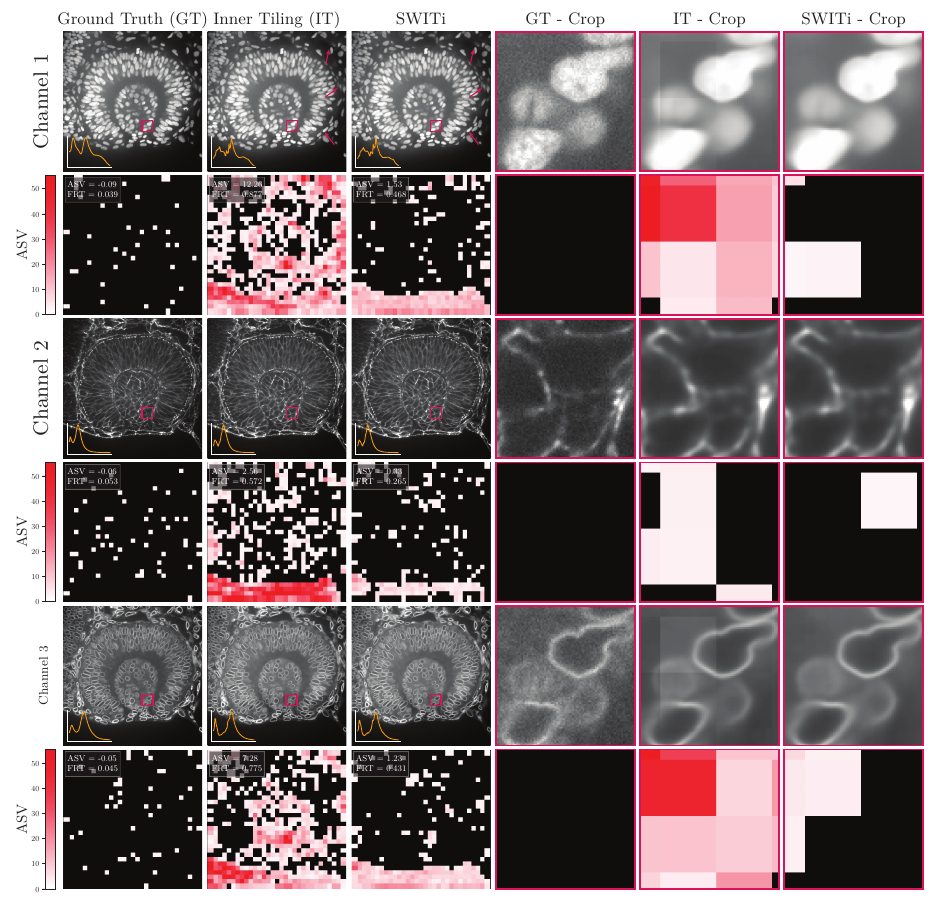}
    \caption{
        \textbf{Qualitative comparison and per-tile seam detection on CBG-Z18.}
        The organization of this figure follows~\cref{fig:3_quali_Pavia}; check its caption for a detailed description. 
        Magenta arrows have been added to highlight regions with higher concentrations of tiling artifacts.
        Inner tiling seams are visible in Channels~1 and~3, as seen in the zoomed-in crops, and far less evident in Channel~2.
        The crops show that SWITi attenuates these artifacts throughout Channels~1 and~3, though it does not remove them entirely: a dimmer residual tiling remains visible at SWITi's stride.
        The test confirms this quantitatively: it triggers detections throughout Channels~1 and~3, whereas in Channel~2, detections are weaker and less certain, with ASV closer to zero; in all channels, SWITi substantially reduces the number of detections.
    }
    \label{fig:suppl_quali_CBGZ18}
\end{figure}

\begin{figure}[t]
    \centering
    \includegraphics[width=\linewidth]{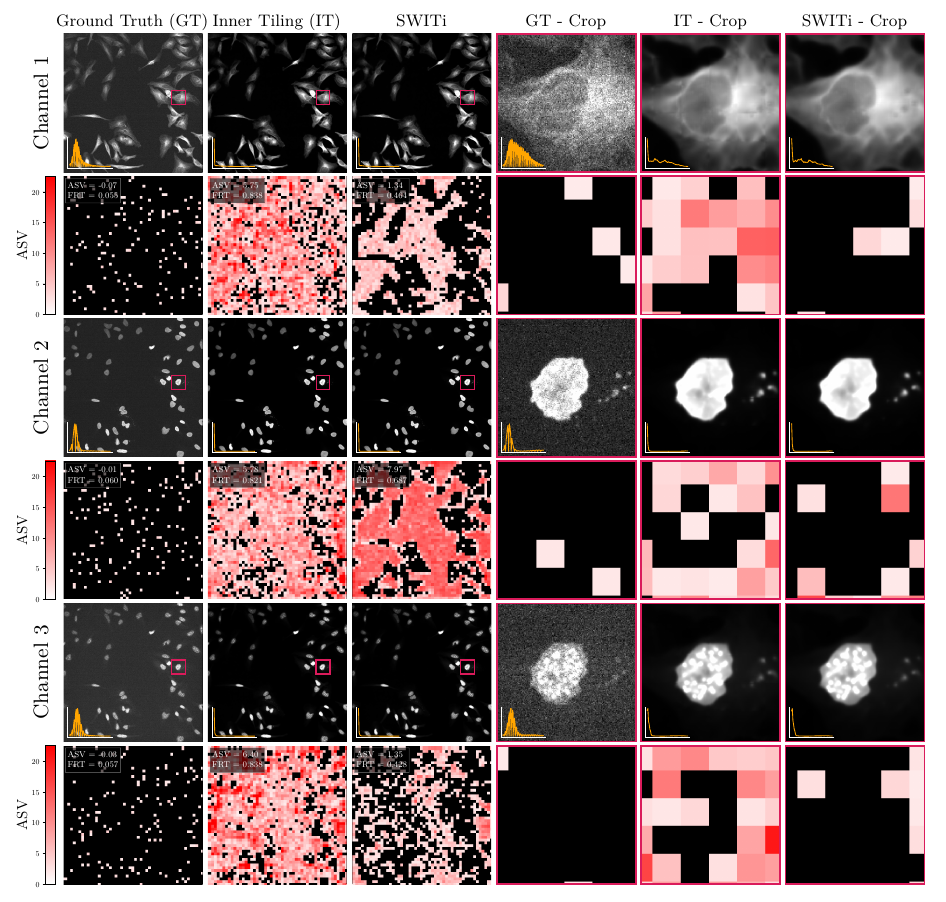}
    \caption{
        \textbf{Qualitative comparison and per-tile seam detection on HT-LIF24 (5ms).}
        The organization of this figure follows~\cref{fig:3_quali_Pavia}; check its caption for a detailed description. 
        %
        Inner tiling seams are visible in the zoomed-in crops, most prominently in Channel~1, and are visually removed in the SWITi reconstructions.
        The test confirms this: it triggers detections throughout the inner tiling predictions.
        For SWITi, FRT and ASV are particularly high for Channel~2; however, most detections are triggered in background areas.
        This is expected, as in flat areas (like the background), even a faint tiling artifact is sufficient to discriminate seam from control gradients.
    }
    \label{fig:suppl_quali_HT5ms}
\end{figure}

\begin{figure}[t]
    \centering
    \includegraphics[width=\linewidth]{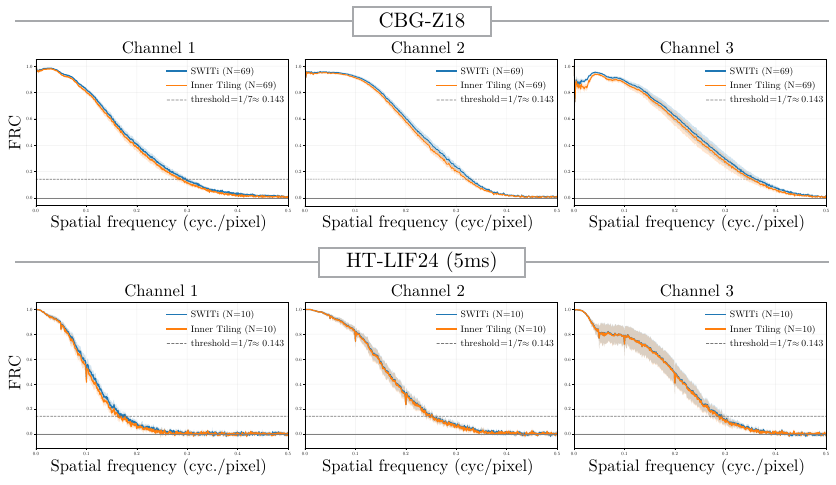}
    \caption{
        \textbf{Fourier Ring Correlation (FRC) curves on CBG-Z18 and HT-LIF24 (5\,ms).}
        Each row shows one dataset, with one panel per split channel; within each panel, FRC between the prediction and the ground-truth reference is plotted against spatial frequency for inner tiling (orange) and SWITi (blue), with shaded bands denoting the standard error over images ($N$ as indicated in each panel).
        The dashed line marks the fixed $1/7 \approx 0.143$ threshold, whose crossing defines the resolution cutoff (higher is better).
        For these datasets, the FRC curve for SWITi is consistently above or on par with the inner tiling curve; still, the curves are closer together than for PaviaATN~(\cref{fig:4_FRC_Pavia}). 
        This is consistent with the fact that seams are generally milder than in PaviaATN, so the improvement from SWITi is lower.
        In any case, where the curves are visually indistinguishable, the cutoff frequencies reported in Table~\ref{tab:frc} still place SWITi at or above inner tiling in nearly every channel, confirming a moderate, but consistent advantage.
    }
    \label{fig:suppl_FRC_others}
\end{figure}

For completeness, here we collect the results that could not be accommodated in the main text.
Global reconstruction metrics (Range-Invariant PSNR, MS-SSIM, LPIPS, and Pearson correlation) for all datasets, per channel and averaged, are reported in~\cref{tab:global_results}.
Qualitative comparisons between inner tiling and SWITi, showing the stitched prediction on a representative image alongside per-tile artifact heatmaps, are provided for CBG-Z18 and HT-LIF24 in~\cref{fig:suppl_quali_CBGZ18,fig:suppl_quali_HT5ms}, respectively.
The corresponding Fourier Ring Correlation curves against the ground-truth reference are shown, again for CBG-Z18 and HT-LIF24, in~\cref{fig:suppl_FRC_others}.
\sloppy

\subsection{Calibration of the Permutation Test}
\label{sec:calibration}

\begin{table}[t!]
\centering
\caption{
\textbf{Calibration of the per-tile gradient-based permutation test on ground truth (GT) images.}
We report the FRT and ASV scores averaged across GT test images for the different datasets, along with the corresponding standard errors (except for CBG-Z18, whose test set comprises a single volume).
A well-calibrated test should yield $FRT\!\approx\!\alpha$ $(\alpha\!=\!0.05)$, and $ASV\!\approx\!0$.
Across datasets, the FRT score correctly stays in the range $0.05$--$0.10$; the mild
over-rejection may reflect residual within-tile pixel dependence.
Calibration of the permutation test is confirmed by the values attained by ASV, which sit closer to zero in all the considered cases.
}
\label{tab:permutation_calibration}
\setlength{\tabcolsep}{4pt}
\begin{tabular}{ll cc}
\toprule
Dataset & Channel & FRT & ASV \\
\midrule
\multirow{2}{*}{PaviaATN}
  & Ch.\,1 & $0.057 \pm 0.001$ & $-0.04 \pm 0.01$ \\
  & Ch.\,2 & $0.058 \pm 0.001$ & $-0.02 \pm 0.00$ \\
\midrule
\multirow{3}{*}{CBG-Z18}
  & Ch.\,1 & $0.098$ & $0.01$ \\
  & Ch.\,2 & $0.067$ & $-0.09$ \\
  & Ch.\,3 & $0.083$ & $-0.07$ \\
\midrule
\multirow{3}{*}{HT-LIF24 (5\,ms)}
  & Ch.\,1 & $0.085 \pm 0.002$ & $-0.29 \pm 0.00$ \\
  & Ch.\,2 & $0.084 \pm 0.002$ & $-0.29 \pm 0.01$ \\
  & Ch.\,3 & $0.087 \pm 0.002$ & $-0.29 \pm 0.01$ \\
\bottomrule
\end{tabular}
\end{table}

The per-tile permutation test should reject the null hypothesis of artifact-free stitching only when an across-seam discontinuity is actually present.
Since the test is non-parametric and its decision is governed by a nominal significance level $\alpha$, it must be calibrated: applied to a seam-free image, the empirical fraction of rejected tiles should match $\alpha$, and the per-tile $z$-scores should have a mean of zero.
Calibration is a crucial step, because otherwise the test would conflate artifacts with the baseline rejection rate (due to random chance).

In the proposed permutation test, the parameter we can tweak for calibration is the block size $B$.
Recall that we permute contiguous blocks of $B$ gradients rather than individual values, because adjacent gradients are spatially correlated through shared image content and, therefore, the exchangeability assumption of the permutation test breaks at the pixel level.
The block size determines the granularity at which this correlation is preserved since the gradients within a block remain together after permutation.
Preserving the sets of correlated gradients keeps the effective number of independent units small, so a random relabeling can still produce a seam/control split as extreme as the observed one; the observed statistic is therefore treated as an ordinary outcome under the null hypothesis, hence making the test more conservative.
When $B$ is small relative to the gradient correlation length, permutation instead scatters correlated pixel gradients across both groups, making every relabeling more likely to produce a uniform split.
Extreme configurations then become rare, so the null distribution of the statistic collapses toward zero and $\tau_{\text{obs}}$, which still carries the real, correlation-induced discrepancy, appears artificially extreme.
This yields systematically small $p$-values and hence over-rejection (an \textit{anti-conservative} test).
On the other hand, further increasing $B$ beyond the correlation length reduces the number of exchangeable units, thereby coarsening the null and lowering the power to detect real artifacts.
The block size thus trades calibration against power.

In our experiments, we selected the smallest $B$ that restores the nominal rejection rate on artifact-free data, which is $B=3$.
In practice, we calibrated $B$ on the PaviaATN dataset, tuning $B$ so that the test applied to the ground truth (artifact-free) images rejects at approximately $\alpha$.
\Cref{tab:permutation_calibration} reports the resulting calibration check across all datasets: median $z$-scores stay close to zero and the rejected-tile fraction remains within $0.05$--$0.10$.
The mild residual over-rejection above $\alpha = 0.05$ reflects within-tile pixel dependence not fully absorbed at the chosen block size, and is most pronounced in the low-SNR HT-LIF24 (5\,ms) data.
As these discrepancies are small and comparable across datasets, we retained the block size calibrated on PaviaATN for all of them rather than tuning $B$ per dataset, keeping a single fixed hyperparameter throughout.



\subsection{Ablation Study of Test Statistics for the Permutation Test}
\label{subsec:statistic_ablation}

The per-tile permutation test of~\cref{sec:meth_metric} is agnostic to the two-sample statistic used to compare the across-seam and control gradient distributions: any measure of the discrepancy between the two empirical histograms can be substituted without altering the test's validity. 
In the manuscript, we use the Jensen--Shannon divergence (JS) as our default.
Here, we report the ablation of different divergence statistics and motivate our choice.

In addition to JS, we consider three other candidates: the Kullback--Leibler (KL) divergence, the Kolmogorov--Smirnov (KS) statistic, and the Wasserstein-1 (W1) distance.
The main criteria we used for choosing the best statistic are the following: 
$(i)$~\emph{calibration}: applied to seam-free ground truth, the per-image rejection rate should be as close as possible to the nominal level $\alpha$;
$(ii)$~\emph{sensitivity}: the statistic must respond strongly in the presence of tiling artifacts in datasets where predictions exhibit evident seams.
$(iii)$~\emph{analytical and numerical properties}: the statistic must possess nice analytical properties, and must be numerically robust and quick to compute.

We perform this ablation study on the PaviaATN dataset.
For the sensitivity analysis, we consider the \emph{global} (\ie, average across channels) FRT and ASV scores on inner tiling predictions, since inner tiling exhibits the most evident artifacts and, hence, provide a more reliable validation.
Results are reported in~\cref{tab:statistic_ablation}.

Regarding \emph{calibration}, we observe that for both KL and JS the FRT over GT images sits at the nominal level, whereas KS and W1 tend to over-reject ($FRT > 0.05$).
For this reason, we can already exclude the latter two statistics.

As far as \emph{sensitivity} is concerned, as FRT does not vary much across statistics, we focus on the ASV score.
We observe that, while all ASV scores for the GT are reasonably close to zero, KL and JS attain the two largest scores on inner tiling predictions (\ie, they are the most sensitive statistics). 
Therefore, the two statistics are also the most suitable choice in this case.

To decide between the two remaining candidates (KL and JS), we consider their analytical and numerical properties.
The Jensen--Shannon divergence is by far the most favorable choice in this regard, as it is bounded (by $\log 2$) and, hence, it cannot diverge to $+\infty$ like the Kullback-Leibler divergence for non-overlapping discrete empirical distributions.

Combining all the observations above, we consequently adopt JS as the default statistic for the per-tile gradient-based permutation test.

\begin{table}[t]
\centering
\caption{
    \textbf{Ablation of the two-sample statistic used in the per-tile permutation test}, on PaviaATN at $M{=}64$ and control-strip width $N{=}2$. 
    We report global FRT and ASV on ground truth (GT) images and inner tiling predictions.
    On GT images, a desirable statistic should provide $\text{FRT}{\approx}0.05$ and $\text{ASV}{\approx}0$. 
    On the other hand, on inner tiling predictions, the statistic should provide $\text{FRT}{\approx}1$ and a large value of $\text{ASV}$, reflecting the pervasive and strong tiling artifacts of PaviaATN dataset.
}
\label{tab:statistic_ablation}
\setlength{\tabcolsep}{6pt}
\begin{tabular}{lcccc}
\toprule
& \multicolumn{2}{c}{global FRT}
& \multicolumn{2}{c}{global ASV} \\
\cmidrule(lr){2-3}
\cmidrule(lr){4-5}
Statistic & Inner Tiling & GT & Inner Tiling & GT  \\
\midrule
\textbf{KL} & 0.98 & 0.050 & 51.2 & 0.010 \\
\textbf{JS} & 0.99 & 0.051 & 29.5 & 0.005 \\
\textbf{KS} & 0.97 & 0.061 & 10.2 & 0.064 \\
\textbf{W1} & 0.98 & 0.066 & 18.7 & 0.078 \\
\bottomrule
\end{tabular}
\end{table}

\subsection{Stochastic versus Deterministic Origin of Residual Seams}
\label{subsec:seams_deterministic}

Tiled inference in posterior models produces tiling artifacts through two compounding mechanisms (\cref{sec:intro}): a deterministic one, related to the incomplete receptive field, and a stochastic one, related to independently sampled latent draws across neighboring tiles.  
In this section, we investigate the relevance of each mechanism in causing tiling artifacts.
We exploit the fact that MMSE estimation is a direct way to mitigate the stochastic source of tiling artifacts: by taking infinitely many samples of a given pixel from its posterior distribution, we should, in principle, remove discrepancies between inner regions, as what we obtain is the mean of that posterior distribution.
In practice, sampling infinitely is impossible;
Nevertheless, we can still assume that the more samples are drawn from the posterior to obtain the MMSE estimate, the closer we get to the true posterior mean, and hence the more the stochastic source of tiling artifacts is attenuated.

To experimentally validate this claim, we compute inner tiling predictions with different MMSE sample counts, namely $M \in \{1, 2, 8, 16, 64, 256\}$ on the PaviaATN dataset.
We assume that $M=256$ is sufficient to provide an approximation of infinite-sampling.
To measure the extent of tiling artifacts, we track the global (\ie, average over channels) ASV score.
The experiment monitors two aspects: 
\begin{enumerate}[itemsep=2pt, topsep=2pt]
    \item~A reduction of global ASV as $M$ increases, confirms that the stochastic mechanism indeed plays a role in the generation of tiling artifacts; if it were not the case, and the deterministic source were the only source, then increasing $M$ would not have any effect.
    \item~The relative drop of ASV between $M=1$ and $M=256$ provides an estimate of the contribution of the stochastic source to tiling artifacts.
\end{enumerate}

\begin{figure}[t]
\centering
\includegraphics[width=0.72\linewidth]{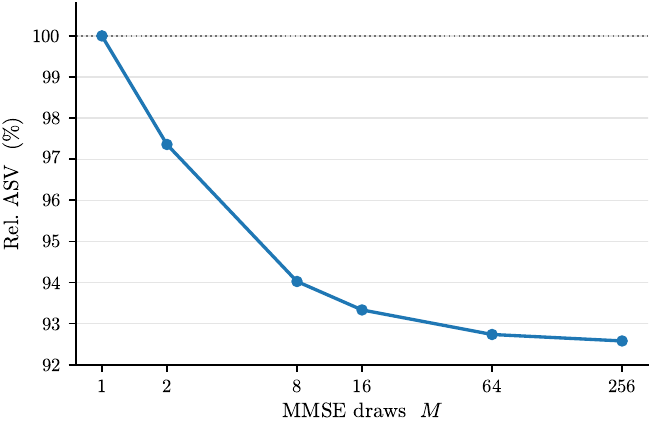}
\caption{
    \textbf{Relative ASV score for increasing MMSE sampling count on PaviaATN.}
    Relative global artifact severity (Rel.\ ASV), as a percentage of its single-draw ($M=1$) value, is plotted against the number of MMSE samples, $M$.
    The drop of ASV as $M$ increases indicates a non-negligible stochastic source of tiling artifacts for this dataset.
    However, at $M=256$, the artifacts still retain approximately $92.5\%$ of the measured severity, meaning that the largest source of tiling artifacts is the deterministic one associated with receptive field truncation.
}
\label{fig:seam_vs_mmse}
\end{figure}

Results reported in~\cref{fig:seam_vs_mmse} show a moderate, but steady decrease of ASV as $M$ increases, confirming a non-negligible contribution to tiling artifacts from the stochastic component.
The overall drop in relative ASV from $M=1$ to $256$ amounts to roughly $7.5\%$.
This means that the largest source of tiling artifacts for Micro$\mathbb{S}plit$, and specifically for the PaviaATN dataset, is the deterministic one.

Finally, we observe that the ASV curve plateaus after $M=64$.
This implies two things: that $M=64$ is a robust choice that minimizes the stochastic component of tiling artifacts at the lowest computational cost, and that $M=256$ is a good approximation of the infinite-sample limit.


\end{document}